\documentclass[10pt,twocolumn,letterpaper]{article}

\usepackage{iccv}
\usepackage{times}
\usepackage{epsfig}
\usepackage{graphicx}
\usepackage{amsmath}
\usepackage{amssymb}
\usepackage{subfigure}
\usepackage{booktabs} 
\usepackage{enumitem}
\usepackage{multirow}

\usepackage{mathtools}
\usepackage{amsthm}
\usepackage{bm}
\usepackage{xspace}
\usepackage{float}

\usepackage[pagebackref=true,breaklinks=true,letterpaper=true,colorlinks,bookmarks=false]{hyperref}

\usepackage[textsize=tiny]{todonotes}

\makeatletter
\DeclareRobustCommand\onedot{\futurelet\@let@token\@onedot}
\def\@onedot{\ifx\@let@token.\else.\null\fi\xspace}

\makeatother

\iccvfinalcopy 

\def\iccvPaperID{16} 

\ificcvfinal\fi

\newcommand{\methodlong}{Historically Guided Domain Adaptation\xspace}
\newcommand{\methodshort}{Hist-DA\xspace}
\newcommand\mypara[1]{\vspace{1.mm}\noindent\textbf{#1}}

\newcommand{\lyft}{Lyft\xspace}
\newcommand{\ith}{Ithaca-365\xspace}

\begin{document}

\title{Unsupervised Domain Adaptation for Self-Driving from\\Past Traversal Features}

\author{
Travis Zhang\thanks{Denotes equal contribution.} \thanks{Correspondences could be directed to \texttt{tz98@cornell.edu}} $^{,1}$\hspace{10pt}
Katie Luo\footnotemark[1] $^{,1}$\hspace{10pt}
Cheng Perng Phoo$^{1}$\hspace{10pt}
Yurong You$^{1}$\hspace{10pt} \\
Wei-Lun Chao$^{2}$ \hspace{10pt}
Bharath Hariharan$^{1}$\hspace{10pt}
Mark Campbell$^{1}$\hspace{10pt}
Kilian Q. Weinberger$^{1}$\\ 
$^1$Cornell University\hspace{14pt}$^2$The Ohio State University
}


\maketitle
\ificcvfinal\thispagestyle{empty}\fi

\begin{abstract}
The rapid development of 3D object detection systems for self-driving cars has significantly improved accuracy. However, these systems struggle to generalize across diverse driving environments, which can lead to safety-critical failures in detecting traffic participants. 
To address this, we propose a method that utilizes unlabeled repeated traversals of multiple locations to adapt object detectors to new driving environments. By incorporating statistics computed from repeated LiDAR scans, we guide the adaptation process effectively. Our approach enhances LiDAR-based detection models using spatial quantized historical features and introduces a lightweight regression head to leverage the statistics for feature regularization. Additionally, we leverage the statistics for a novel self-training process to stabilize the training. The framework is detector model-agnostic and experiments on real-world datasets demonstrate significant improvements, achieving up to a 20-point performance gain, especially in detecting pedestrians and distant objects. Code is available at \url{https://github.com/zhangtravis/Hist-DA}.
\end{abstract}

\section{Introduction}
\label{sec:intro}

Self-driving cars need to detect objects like cars and pedestrians and localize them in 3D to drive safely. 3D object detection systems have advanced rapidly in accuracy, but still fail to generalize across the extremely diverse domains where vehicles are deployed:
A perception system trained in sunny California may never have seen snow-covered cars, and may fail to detect these cars with disastrous consequences.
Unfortunately, we cannot afford to separately annotate training data for every location a car might be driven in. We therefore need ways of adapting 3D perception systems to new driving environments without labeled training data. This is the problem of \emph{unsupervised domain adaptation}, where the object detector must be adapted to a new \emph{target} domain where only unlabeled data is available. In this work, we explore 3D object detection from LiDAR data and how to best adapt it to a set of diverse, real-world scenarios.

Different from prior works in unsupervised domain adaptation,
we follow~\cite{you2022unsupervised} to include the assumption that unlabeled repeated traversals of the same locations are available to the adaptation algorithm.
As discussed in the prior works, such assumptions are highly realistic: for example, roads and intersections are usually visited many times by many vehicles.
Prior work has shown that the additional information from repeated traversals helps 3D detection in the same domain~\cite{you2022hindsight,you2022learning}, and helps the perception models adapt to a new domain~\cite{you2022unsupervised}.

However, it is not readily obvious how to best utilize the repeated traversals.
Rode-DA~\cite{you2022unsupervised} uses P2-score, which is a statistic computed from repeated LiDAR scans characterizing the persistence of different areas of the 3D scene, to correct the false positive detections in self-training and better supervise the model.
We argue that this method has not fully exploited the information from the P2-score and it can be used in a more principled way to guide the adaptation process.



Our key insight is that the P2-score is a perfect signal
to \emph{regularize} the feature in the detection training.
Our full method is based on Hindsight~\cite{you2022hindsight}.
Hindsight enhances LiDAR-based 3D object detection models
with the spatial quantized historical (SQuaSH) features computed from repeated past traversals.
The authors show that Hindsight can greatly improve the detection performance
when tested within similar areas.
However, the SQuaSH features do not guarantee to be invariant
across different domains, resulting in limited performance gain.
To prevent the SQuaSH features overfit to the training domain,
we propose to add the P2-score prediction task as an auxiliary task while training the SQuaSH featurizer.
Observing LiDAR points within each voxel sharing similar P2-score,
we apply an extra light-weighted regression head after the SQuaSH feature,
and train the head with simple P2 regression task.
The regression head is only used in training and does not introduce latency overhead during testing.

Pairing with the typical self-training technique in domain adapation,
we validate our method on two large, real-world datasets: Ithaca365~\cite{carlos2022ithaca365} and Lyft~\cite{lyft2019}, as well as a suite of representative object detectors. Our method, which we term \methodlong (\methodshort), can achieve up to 20 points in improvement, most notably in difficult cases such as detecting pedestrians and far away objects. Furthermore, our method requires very little tuning to achieve strong performance for 3D object detection. Concretely, our contributions are as follows:

\begin{itemize}[itemsep=.3em]
    \item Our methodology identifies a strong source of information with a high learning signal to improve self-supervised adaptation.
    \item We designed a model-agnostic adaptation framework to leverage repeated traversals effectively.
    \item We empirically validated our approach on two real-world datasets and show through ablation studies that \methodshort is robust and generalizable.
\end{itemize}
\section{Related Works}
\label{sec:related}

\begin{figure*}[t]
    \centering
    \includegraphics[width=\textwidth]{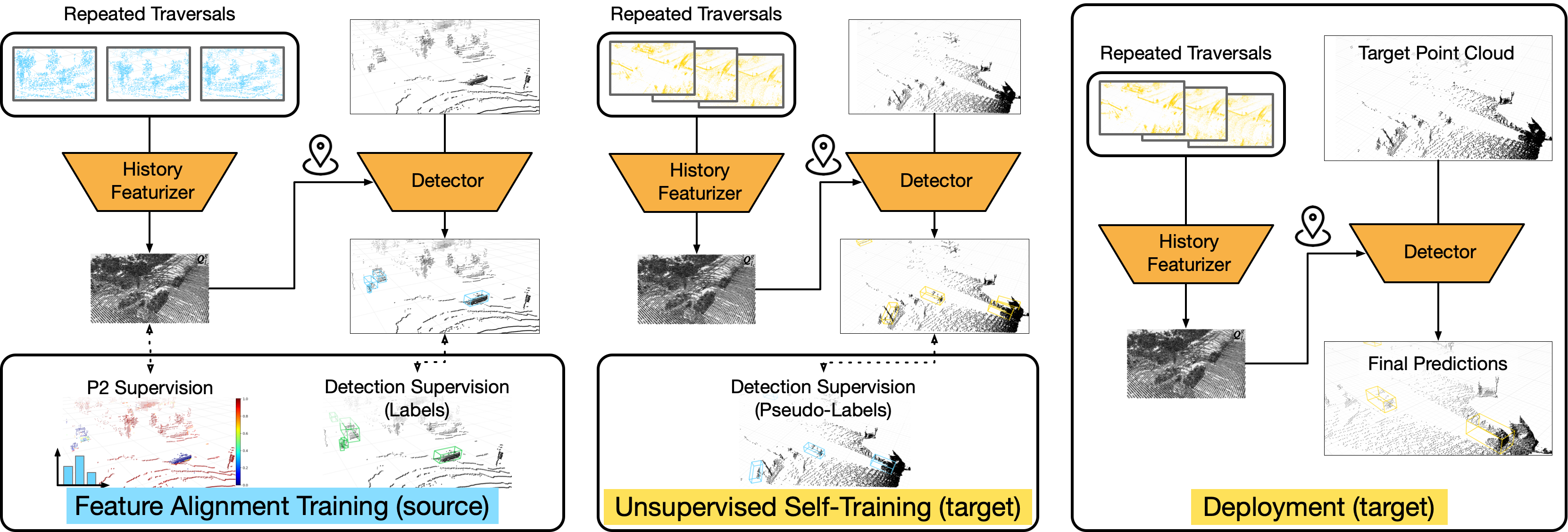}

    \vspace{0.5\baselineskip}
    \caption{\textbf{Method diagram of the adaptation process.} The method is divided into source domain training, target domain unsupervised training, and finally deployment on the target domain. The repeated traversals from the source domain are colored in blue, and those from the target domain are colored in yellow. Best viewed in color.}
    \label{fig:methodology}
    \vspace{-1.2\baselineskip}
\end{figure*}


\mypara{Past Traversals in Autonomous Driving} Human drivers often drives through the same locations repeatedly, thus it is natural to assume that the (unlabeled) data collected for training perception systems for self-driving vehicle contains repeated traversals of different locations. Past works have leveraged this property to enhance the perception of autonomous vehicles. These include self-supervising 2D representation for visual odometry \cite{barnesephemerality}, uncovering mobile objects in LiDAR in an unsupervised manner \cite{you2022learning}, etc. The line of work that is directly related to ours is Hindsight \cite{you2022hindsight} where the authors proposed to learn additional feature descriptors for each point in a LiDAR point cloud from the unlabeled past traversals for better downstream 3D object detection. Hindsight is simple and effective and would work with any downstream 3D detectors that consumes 3D LiDAR point cloud. In this work, we seek to adapt this family of detectors when deploying to a new domain where unlabeled past traversals are available. 

\mypara{Unsupervised Domain Adaptation (UDA) for 3D Object Detection.} Adapting 3D object detectors to new domains where no labels are available is crucial to deployment of self-driving vehicles. The key to UDA is to understand the domain differences that the detector would encounter during deployment. SN discovers that car sizes could be a source of domain differences and propose to normalize the car sizes when training the detectors on the source domain \cite{wang2020train}; SPG identifies point cloud density as one potential source of differences and propose to fill in point clouds during deployment \cite{xu2021spg}. Though these methods have shown remarkable progress in the problem, they all target specific domain differences, which is not feasible in all cases. One way to characterize domain differences is through the use of unlabeled data. Along this vein, ST3D \cite{yang2021st3d} and ST3D++ \cite{yang2019std} adopt conventional self-training approaches with improved filtering mechanism to stabilize adaptation whereas MLC-Net \cite{luo2021MLC-Net} achieves domain alignment via enforcing consistency between a source detector and its exponential moving average on the unlabeled data. Though these methods\cite{luo2021MLC-Net,wang2020train,yang2021st3d} are effective, they mostly assume that all the unlabeled data are i.i.d. which ignores other potential signals that could be inherent in the unlabeled data such as temporal signals \cite{you2022unsupervised} that are potentially useful in adaptation. In this work, we explore using unlabeled past traversals for domain adaptation. As shown in \cite{you2022unsupervised}, these correlated data contains potent signals for aiding adaptation. However, crucially different from \cite{you2022unsupervised}, we focus on adapting Hindsight --- a family of models that uses past traversals during inference time.





\section{\methodlong}
\label{sec:method}



We attempt to adapt a 3D object detector to a target domain using unlabeled data. Different from typical adaptation setup~\cite{wang2020train, yang2021st3d}, we assume the autonomous driving system has access to multiple traversals of the same driving scenes and accurate localization information, both in the source and target domain.
In section \ref{sec:adaptation_setup}, we will clearly lay out the adaptation setup. Then, we will discuss relevant background information to clarify our proposed methodology in section \ref{sec:background}.
Our key insight is to leverage P2-score information from repeated traversals to adapt the detector's point features from one domain to another ---akin to feature-alignment works done in the 2D space--- as well as self-training to ensure stable predictions. We discuss the relevant adaptation strategies in section \ref{sec:adapt_strat}. Our overall method is shown in Figure~\ref{fig:methodology}.

\subsection{Unsupervised Domain Adaptation with Repeated Traversals}
\label{sec:adaptation_setup}
Our goal is to adapt a LiDAR-based detector using repeated traversals of unlabeled point clouds  $\{\bm{P}_i^t\}$  and the associated global localization $\{G_i^t\}$ from the target domain, for the $i$-th frame in traversal $t$. 
We assume that the source domain also has access to multiple repeated traversals, as well as the bounding box labels $\bm{b}_c$ associated with training point clouds $\bm{P}_c$, for the $c$-th training frame.
To characterize these historical traversals, we combine the point clouds for a single traversal to create a dense point cloud in the same way as \cite{you2022hindsight}. Specifically, for a single traversal in a domain that consists of a sequence of point clouds, we transform each point cloud into a fixed global coordinate system.
Then, for a location $l$ in a single frame $i$ within traversal $t$ every $m$ meters along the road, the point clouds from a range $[-H_m, H_m]$ are combined to produce a dense point cloud $\bm{D}_{l}^t=\bigcup_{G_i^t\in[l-H_m, l+H_m]}\{\bm{P}^i_{t}\}$ (with a slight abuse of notation as we use $G_i^t$ additionally for the location $i$ was captured). 


\subsection{Background}
\label{sec:background} 


\textbf{Persistency Prior Score from multiple traversals.}
Our goal is to exploit the inherent information from the unlabeled repeated traversals for adaptation. One source of information we can retrieve is the Persistency Prior (P2) score, that was introduced in \cite{you2022learning}. To recap, P2-score uses entropy-based measures to quantify how persistent a single LiDAR point cloud is across multiple traversals. It is calculated using the set of dense point clouds $\{\bm{D}_{l}^t\}_{t=1}^T$, for $T\geq 1$ traversals of a location. For a given 3D point $\bm{q}$ around location $l$, we first count the number of neighboring points around $\bm{q}$ within a certain radius $r$ in each $\bm{D}_{l}^t$:
\begin{equation}
    N_t(\bm{q}) = \big|\{\bm{p}_i ; ||\bm{p}_i - \bm{q}||_2 < r, \bm{p}_i \in \bm{D}^t_{l}\}\big| 
\end{equation}
We can then normalize the neighbor count $N_t(\bm{q})$ across traversals $t \in \{1,...T\}$ into a categorical probability:
\begin{equation}
    P(t;\bm{q}) = \frac{N_t(\bm{q})}{\sum_{t'=1}^T N_{t'}(\bm{q})}
\end{equation}
Using $ P(t;\bm{q})$, we can then compute the P2-score $\tau(\bm{q})$ the same way as \cite{you2022learning}:
\begin{equation}
    \tau(\bm{q}) = \begin{cases} 
      0 & \text{if } N_t(\bm{q}) = 0 \text{ } \forall{t}; \\
      \frac{H(P(t;\bm{q}))}{\log(T)} & \text{otherwise}
   \end{cases}
\end{equation}
where $H$ is the information entropy. Intuitively, a higher P2-score corresponds to a more persistent background, while a lower P2-score corresponds to a mobile foreground object.
This value is a statistic that can be calculated from the repeated traversals, and as a result, it's natural for us to use an architecture that leverages these data. 
One such candidate is Hindsight. 

\textbf{Hindsight.}
Hindsight \cite{you2022hindsight} is an end-to-end featurizer intended to extract contextual information from repeated past traversals of the same location. The authors proposed an easy-to-query data structure used to endow the current point cloud with information from past traversals to improve 3D detection. 

Given the dense point cloud $\bm{D}_{l}^t$, Hindsight encodes it using a spatial featurizer that results in a spatially-quantized feature tensor $\bm{Q}_{l}^t$. This can be applied to the $T$ tensors, one for each traversal in location $l$, which is then aggregated into a single tensor $\bm{Q}_l^g$, deemed SQuaSH, using a per-voxel aggregation function $f_{agg}$:
\begin{equation}
    \bm{Q}_l^g = f_{agg}(\bm{Q}_{l}^1,...,\bm{Q}_{l}^{T})
\end{equation}
Once deployed, if the self-driving car captures a new scan $\bm{P}_c$ at a new location $G_c$ and the SQuaSH feature at this location is $\bm{Q}^g_{l_c}$, Hindsight endows $\bm{P}_c$ by querying the features $\bm{Q}^g_{l_c}$ around it.
In the work \cite{you2022hindsight}, the SQuaSH featurizer is trained concurrently with the object detector, using the detection loss as a signal for gradient updates.







\begin{table*}[h]
\tabcolsep 4pt
\centering
\begin{tabular}{@{}lccccccccc@{}}
\toprule
                        & \multicolumn{4}{c}{Car}                                                                   &                      & \multicolumn{4}{c}{Pedestrian}                                                            \\ \cmidrule(lr){2-5} \cmidrule(l){7-10} 
Method                  & 0-30                 & 30-50                & 50-80                & 0-80                 &                      & 0-30                 & 30-50                & 50-80                & 0-80                 \\ \midrule
No Adapt/ No HS         & 42.19                & 12.66                & 0.95                 & 18.54                &                      & 40.74                & 18.32                & 0.42                 & 21.18                \\
ST3D   &  61.63	& 38.70& 	4.73 & 	35.89	& & 44.37	& 26.94 & 	0.00 &	24.97  \\       
Rote-DA                 & \textbf{62.85}	 & \textbf{41.88}	& 15.07 & 	41.32	& & 48.76 & 	32.61	& 1.21 &	30.59                \\ \midrule
No Adapt + HS & 41.88	& 29.31	& 16.40	& 30.29	& & 51.16	& 26.41	& 5.80	& 29.99 \\
Hist-DA (Ours)         							           & \multicolumn{1}{c}{58.44} & \multicolumn{1}{c}{40.03} & \multicolumn{1}{c}{\textbf{25.26}} & \multicolumn{1}{c}{\textbf{42.82}} &                      & \multicolumn{1}{c}{\textbf{60.72}} & \multicolumn{1}{c}{\textbf{48.58}} & \multicolumn{1}{c}{\textbf{21.42}} & \multicolumn{1}{c}{\textbf{48.48}} \\
{\color[HTML]{9B9B9B} Oracle (in domain)} & 							\multicolumn{1}{c}{\color[HTML]{9B9B9B} 73.38} & \multicolumn{1}{c}{\color[HTML]{9B9B9B} 56.19} & \multicolumn{1}{c}{\color[HTML]{9B9B9B} 39.08} & \multicolumn{1}{c}{\color[HTML]{9B9B9B} 57.10} &                      & \multicolumn{1}{c}{\color[HTML]{9B9B9B} 55.39} & \multicolumn{1}{c}{\color[HTML]{9B9B9B} 37.42} & \multicolumn{1}{c}{\color[HTML]{9B9B9B} 14.86} & \multicolumn{1}{c}{\color[HTML]{9B9B9B} 40.37 } \\ \bottomrule
\end{tabular}

\vspace{4px}
\caption{Results of adapting a detector from Lyft to Ithaca365. Metrics are reported on nuScenes mAP at 1m matching.}
\label{tab:detection-lyfttoith-07bev}
\end{table*}
\begin{table*}[ht]
\tabcolsep 4pt
\centering
\begin{tabular}{@{}lccccccccc@{}}
\toprule
                        & \multicolumn{4}{c}{Car}     & \multicolumn{1}{l}{} & \multicolumn{4}{c}{Pedestrian} \\ \cmidrule(lr){2-5} \cmidrule(l){7-10} 
Method                  & 0-30 & 30-50 & 50-80 & 0-80 & \multicolumn{1}{l}{} & 0-30  & 30-50  & 50-80  & 0-80 \\ \midrule
No Adapt/ No HS         & 59.0 & 40.9  & 25.8  & 45.4 &                      & 16.7  & 8.2    & 0.2    & 6.7  \\
ST3D                    &  \textbf{71.8} &	\textbf{52.1} &	30.4 &	\textbf{55.7} & &	 -- &	 -- &	 --	 & --      \\
Rote-DA                 & 54.3 &	31.9 &	14.9 &	35.7 & &	\textbf{29.6} &	\textbf{34.4} &	4.1 &	\textbf{22.0}  \\ \midrule
\methodshort (Ours)                    & 62.6 &	49.2 &	\textbf{34.9} &	51.8 &	& 25.6 &	26.5 & 	\textbf{7.9} &	16.7 \\ 
{\color[HTML]{9B9B9B} Oracle (in domain)} &  {\color[HTML]{9B9B9B} 69.1} &	{\color[HTML]{9B9B9B} 71.5} &	{\color[HTML]{9B9B9B} 49.0} &	{\color[HTML]{9B9B9B} 65.7} &&	{\color[HTML]{9B9B9B} 37.0} &	{ \color[HTML]{9B9B9B} 38.2} &	{\color[HTML]{9B9B9B} 26.3} & {\color[HTML]{9B9B9B} 32.2}     \\ \bottomrule
\end{tabular}

\vspace{4px}
\caption{Results of adapting a detector from Ithaca365 to Lyft. Metrics are reported at 0.7 IoU matching.}
\label{tab:detection-ithtolyft-07bev}
\end{table*}

\subsection{Adaptation Strategy}
\label{sec:adapt_strat}
Our adaptation strategy consists of P2 feature alignment training in the source domain and unsupervised self-training in the target domain.

\paragraph{P2 Feature Alignment Training} Though the computation of P2-score is of high latency and thus it is hard to be applied online, it serves perfectly as an additional signal for offline adaptation algorithm.
With P2-score, we construct a simple self-supervised learning task to adapt the SQuaSH features after deployment.

Consider a point $\bm{q}$ in a point cloud $\bm{P}_c$, we can obtain 1) its corresponding SQuaSH feature $\mathbf{Q}_l^g(\bm{q})$; 2) its corresponding P2-score $\tau(\bm{q})$. Since the SQuaSH feature is computed from the same traversals, it should contain sufficient information to reproduce the corresponding P2-score. 
However, the trained model might suffer from the domain difference, and thus in the target domain, it might not be able to encode sufficient information from the past traversals, including those for the P2 score.
We thus construct a P2 score prediction task for the SQuaSH feature to help the model align the relevant information it extracts in the source domain to invariant information encoded in P2-scores. For each SQuaSH feature $\bm{Q}_l^g(\bm{q})$, we apply a simple MLP to predict the corresponding P2 score,
\begin{equation}
    \hat{\tau}(\bm{q}) = \mathrm{MLP}(\bm{Q}_l^g(\bm{q})).
\end{equation}
We compute the L1 distance between the predicted P2 score and the corresponding P2 score as the alignment loss,
\begin{equation}
    l_{\text{alignment}} = \|\hat{\tau}(\bm{q}) - \tau(\bm{q})\|_1.
\end{equation}
The final objective for the detector training under the source domain consists of the alignment loss $l_{\text{alignment}}$, in addition to the regular detection loss for the detector we are adapting $l_{\text{detection}}$, computed from the predicted bounding boxes $\hat{\bm{b}}$ and the labels $\bm{b}_c$. Our methodology is detector agnostic, and we do not assume the base detector or the detection loss $l_{\text{detection}}$. 

\paragraph{Unsupervised Self-Training}
To stabilize finetuning in the target domain, we apply self-training in the target domain. Similar to works \cite{lee2013pseudo, you2022unsupervised} that showed the effectiveness of self-training, we leverage refined \textit{pseudo-labels} that we generate for adaptation into the target domain.

Given an aligned detector from P2 feature alignment training, we can generate bounding boxes in the target domain. Given a point cloud $\bm{P}_c$ in the target domain, we can obtain bounding boxes $\hat{\bm{b}}$ from the detector. Similar to the source domain, we can compute the P2-score for each point cloud in the target domain, $\tau(\bm{q})$, $\bm{q} \in \bm{P}_c$. To assess the quality of a particular bounding box, we can apply a simple criteria that points within the bounding box cannot be too \textit{persistent}, \ie having P2-scores that are too high. In this work, we filter out bounding boxes that capture points with P2-scores with a 20\textit{th}-percentile larger than 0.7:
\begin{equation}
    \hat{\bm{b}}_{\text{final}} = \large\{ b \in \hat{\bm{b}} | \text{P}_{20}(\{\tau(\bm{q}_j)\}_{j \in b} ) < 0.7 \large\},
\end{equation}
with a slight abuse in notation, we denote $j \in b$ as the $j$-th point that is in bounding box $b$. This gives us the final set of pseudo-labels, $\hat{\bm{b}}_\text{final}$, and we compute the detection loss for the model on the pseudo-labels, and the final objective for the unsupervised training on the target domain is this pseudo-label detection loss, $l_\text{detection}$ computed on $\hat{\bm{b}}_\text{final}$ as the labels.

\label{sec:method_adapt}
\section{Experiments}
\label{sec:experiments}

\mypara{Datasets.}
We experiment with two large-scale autonomous driving datasets: the Lyft Level 5 Perception dataset~\cite{lyft2019} and the Ithaca-365 dataset~\cite{carlos2022ithaca365}. To the best of our knowledge, these are the only two publicly available autonomous driving datasets that have both bounding box annotations and multiple traversals with accurate 6-DoF localization. 
The \lyft dataset is collected in Palo Alto (California, US) and the \ith dataset is collected in Ithaca (New York, US). 
We use the roof LiDAR (40/60-beam in Lyft; 128-beam in Ithaca-365), 
and the global 6-DoF localization with the calibration matrices directly from the raw data.
We simulate adaptation scenarios in both ways: 1) train in \lyft and test in \ith; 2) train in \ith and test in \lyft.

\begin{figure*}
    \centering
    \includegraphics[width=\linewidth]{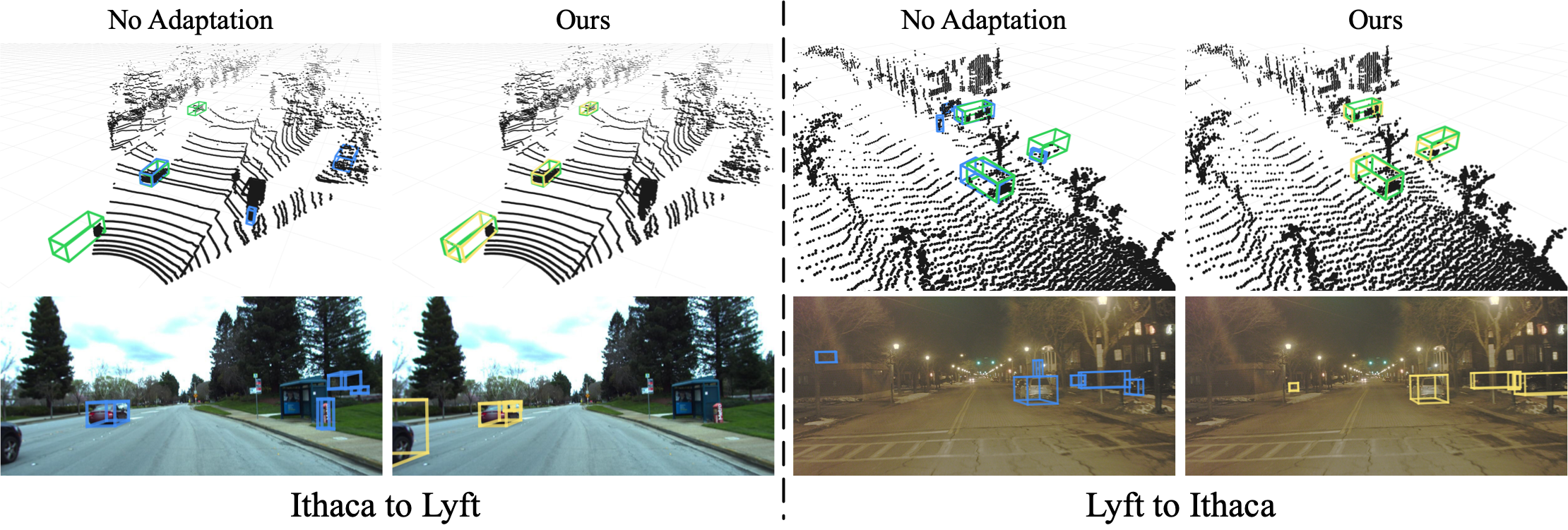}
    \caption{\textbf{Qualitative visualization of adaptation results.} We visualize one example scene (above: LiDAR, below: image, not used for adaptation) from the adaptation results from the \ith $\rightarrow$ \lyft and \lyft $\rightarrow$ \ith datasets. Ground-truth bounding boxes are shown in green, detection boxes of no adaptation and our method are shown in blue and yellow, respectively.  Best viewed in color.}
    \label{fig:qualitative}
\end{figure*}

\mypara{Source 3D Object Detectors.}
In the source domain, we train the default implementation of PointRCNN model with Hindsight \cite{you2022hindsight} using both object detection and P2-score feature alignment training for 60 epochs. We modify the Hindsight model to predict P2-scores from the inputted dense point cloud $\mathbf{S}_l^t$. All models are trained with 4 GPUs (NVIDIA A6000).

It is worth noting that our methodology has the ability to be applied to other 3D object detectors as well, and leave this for future exploration.

\mypara{Evaluation Metrics.}
On the Lyft dataset, we  evaluate object detection performance in a bird's eye view (BEV) and use KITTI \cite{geiger2012we} metrics and conventions for 3D detection. We report average precision (AP) with the intersection over union (IoU) thresholds at 0.7 and 0.5 for Car and Pedestrians. Additionally, these evaluations are evaluated at various depth ranges. Due to space constraints, we report AP$_{BEV}$ at IoU=0.7 for Cars and Pedestrians. On the Ithaca365 dataset, the default match criterion is by the minimum distance to the ground-truth bounding boxes. We report the mean average precision (mAP) with match thresholds of 1-meter for Cars and Pedestrians. 
Since there are too few cyclists in the Ithaca-365 dataset to provide a reasonable performance estimate, we train and evaluate our models only on \textit{Cars} and \textit{Pedestrians}.

\mypara{Adaptation Method Comparisons}
We compare the proposed methodology against the following methods with publicly available code: ST3D \cite{yang2021st3d} and Rote-DA \cite{you2022unsupervised}. 

\subsection{Domain Adaptation Performance Results}
\begin{table*}[h]
\tabcolsep 2pt
\centering
\begin{tabular}{@{}clclccccccccc@{}}
\midrule
Source               &  & \multicolumn{2}{c}{Target}                                       & \multicolumn{4}{c}{Car}     & \multicolumn{1}{l}{} & \multicolumn{4}{c}{Pedestrian}              \\ \cmidrule(lr){5-8} \cmidrule(l){10-13} 
P2 Training          &  & P2 Training          & \multicolumn{1}{c}{Pseudo-Label} & 0-30 & 30-50 & 50-80 & 0-80 & \multicolumn{1}{l}{} & 0-30 & 30-50 & 50-80 & 0-80                 \\ \midrule
\multicolumn{1}{l}{} &  & \multicolumn{1}{l}{} &                                           & 41.88	&29.31	&16.4&	30.29&&	51.16	&26.41	&5.8	&29.99                  \\
$\checkmark$         &  & \multicolumn{1}{l}{} &                                           & 50.57	&34.57&	19.04&	36.39&&	55.9&	38.83&	12.28&	39.88              \\
$\checkmark$         &  & $\checkmark$         &                                           & 37.24&	15.26&	2.85&	18.37&&	41.16&	26.32 &	12.05 &	27.73                 \\
$\checkmark$         &  & \multicolumn{1}{l}{} & \multicolumn{1}{c}{$\checkmark$}          &   \textbf{58.44}	&40.03&	25.26	&42.82	&&60.72 &	\textbf{48.58}	& \textbf{21.42}	& \textbf{48.48} \\
$\checkmark$         &  & $\checkmark$         & \multicolumn{1}{c}{$\checkmark$}          &      58.06 & \textbf{42.34}& \textbf{25.82}& \textbf{43.31} && \textbf{60.87} &  48.42 &  20.41&  47.92 \\ \bottomrule
\end{tabular}

\vspace{4px}
\caption{Ablation results adapting a detector from Lyft to Ithaca365. Metrics are reported on nuScenes mAP at 1m matching.}
\label{tab:ablation-adaptation-05bev}
\end{table*}
In \autoref{tab:detection-lyfttoith-07bev} and \autoref{tab:detection-ithtolyft-07bev}, we show the results on adaptation from a detector trained in the Lyft dataset to the Ithaca365 dataset and vice versa. Based on the tables, we can see that our methodology, despite its simplicity, outperforms all baselines in almost all metrics in both adaptation directions. This goes to show not only that using multiple traversals serves as a strong learning signal for these models, but also that predicting P2-scores as a self-supervised learning task leads to a dramatic improvement. 

\methodshort works especially well in more challenging scenarios, specifically in the pedestrian scenario and with farther distances. Although it performs slightly worse for cars at close ranges, our methodology has a significantly stronger performance for pedestrians and for far away objects. 
The model even outperforms an in-domain detector in all distances for pedestrians by a substantial amount.
Furthermore, due to the simple nature of the single round of self-training in \methodshort, our method is significantly simpler to train than any of the baselines, which require many rounds of self-training. Consequentially, it is significantly simpler to tune and is faster to train. Observe that by adding in Hindsight features (+ HS), we are already able to observe performance gains over the model that doesn't leverage past traversal information. This shows that such historical features already improve adaption and are more robust across domains. By including our method, we are able to achieve the best performance by explicitly bootstrapping in the past traversal statistics in the form of P2-scores.


\subsection{Qualitative Results}
We visualize our adaptation results in \autoref{fig:qualitative}, and compare the detections of \methodshort (in yellow) to detections without adaptation (in blue). Observe that detection results using \methodshort are qualitatively better than those without adaptation, both in the shape, as well as precision and recall. In particular, for smaller actors such as pedestrians and in actors that are further away. The feature alignment training allows for more robust features that generalizes across domains, and the unsupervised self-training allows for stronger adaptation into the new domain.

\subsection{Analysis}
\mypara{Effect of different adaptation components.} We additionally ablate the different components and report our results in \autoref{tab:ablation-adaptation-05bev}. Observe that adding in P2-score training is crucial to the generalizability of the features across domains. Additionally, adding in self-training (``Pseudo-Label") helps stabilize the model training in an unsupervised manner into the new domain. Although using both P2 Training and pseudo-labels and only using P2 Training have similar performances, we noticed that the number of traversals from the source to the target domain can affect the performance of including P2 Training in the target domain. This occurs because P2 scores are inherently derived from repeated traversals, and the number of traversals can affect its accuracy. To be more specific, we observed that going from Ithaca365, which had 20 traversals to Lyft, which had 5 traversals made the performance of both P2 and pseudo-labels worse than using pseudo-labels since the P2 scores derived from the target domain was introducing noise to cause the model's P2 backbone to decrease in accuracy.

\mypara{Effect of historical traversals.} 
We examine the effect of the additional information by including unlabeled, historical traversals. We report our findings in \autoref{tab:ablation-architecture-05bev}. Although directly evaluating on Ithaca365 using a PointRCNN model trained on Lyft has noticeable performances, one can see that including the Hindsight model increases the performance in both cars and pedestrians, with some distances improving by almost two-fold. On the other hand, adapting a PointRCNN model without Hindsight leads to improvements specifically for pedestrians, but performs slightly worse than (-) adapt / HS in cars. Naturally, adding both would significantly improve the performance as shown in the table, with adapting improving the pedestrian performance and the hindsight model improving the car performance. 

\begin{table}[ht]
\tabcolsep 1pt
\centering
\begin{tabular}{@{}llcccccccc@{}}
\toprule
\multirow{2}{*}{\begin{tabular}[c]{@{}l@{}}PRCNN \\ Model\end{tabular}} & \multicolumn{4}{c}{Car}                                                                                     &  & \multicolumn{4}{c}{Pedestrian}                                                                              \\ \cmidrule(lr){2-5} \cmidrule(l){7-10} 
                                                                        & \multicolumn{1}{c}{0-30} & \multicolumn{1}{c}{30-50} & \multicolumn{1}{c}{50-80} & \multicolumn{1}{c}{0-80} &  & \multicolumn{1}{c}{0-30} & \multicolumn{1}{c}{30-50} & \multicolumn{1}{c}{50-80} & \multicolumn{1}{c}{0-80} \\ \midrule
baseline                                                                &                         42.2	& 12.7	 & 0.9 & 	18.5&&	40.7	&18.3 &	0.4&	21.2                       \\ 
(-) adapt, HS                                                            &                          41.9	& 29.3 & 	16.4 & 	30.3 &&	51.2 &	26.4	& 5.8	& 30.0                        \\ 
adapt, (-) HS                                                                &                         54.1	& 22.8	 & 2.0 &	27.0 &&	52.5&	31.2&	1.9 &	32.1                          \\ \midrule
Ours & \textbf{58.4} & \textbf{40.0}	& \textbf{25.3}	&\textbf{42.8}&&	\textbf{60.7}	& \textbf{48.6}&	\textbf{21.4}&	\textbf{48.5} \\
\bottomrule
\end{tabular}

\vspace{4px}
\caption{Ablation results adapting a detector from Lyft to Ithaca365. Metrics reported on nuScenes mAP at 1m matching.}
\label{tab:ablation-architecture-05bev}
\end{table}

\mypara{Robustness of the framework.}
We analyze the robustness of our method to localization error and number of past traversals used in computing the historical features of the model.
Results for localization error are shown in \autoref{tab:robustness-localization-05bev}; \methodshort is robust to minor errors in noise. We additionally report results for robustness under number of past traversals in \autoref{tab:robustness-traversals-05bev}. Observe that performance gain in adaptation can be seen with even two past traversals of an area. Additionally, our method handles higher depths better than other methods as shown in \autoref{tab:detection-lyfttoith-07bev} and \autoref{tab:detection-ithtolyft-07bev}, since P2-score as a self-supervision task acts as a prior over the point clouds and inherently removes static objects that normal object detectors might not catch at higher depths.
\begin{table}[]
\tabcolsep 1pt
\centering
\begin{tabular}{@{}llcccccccc@{}}
\toprule
Loc. Error & \multicolumn{4}{c}{Car}                                                                                     &  & \multicolumn{4}{c}{Pedestrian}                                                                              \\ \cmidrule(lr){2-5} \cmidrule(l){7-10} 
           & \multicolumn{1}{c}{0-30} & \multicolumn{1}{c}{30-50} & \multicolumn{1}{c}{50-80} & \multicolumn{1}{c}{0-80} &  & \multicolumn{1}{c}{0-30} & \multicolumn{1}{c}{30-50} & \multicolumn{1}{c}{50-80} & \multicolumn{1}{c}{0-80} \\ \midrule
baseline   &  42.2	& 12.7	& 0.9 &	18.5	&  & 40.7 & 	18.3	& 0.4	& 21.2                       \\ \midrule
0.1 m      &     \textbf{58.2}	&\textbf{40.8}&	\textbf{24.6}&	\textbf{42.6} & &	\textbf{59.7} &	\textbf{47.8}&	\textbf{21.1}&	\textbf{47.5}                       \\
0.2 m      &   \textbf{58.2}&	40.1	&23.9&	42.4 &&	59.2&	46.5&	19.4&	46.4                         \\
0.3 m      &     57.3	&40.5	&23.0&	41.8	&&57.4&	44.5&	16.7&	44.2                         \\
0.4 m      &      57.0&	38.8	&21.6&	40.7&&	56.0&	41.2&	14.3&	41.9                          \\
0.5 m      &  56.9	&38.4	&19.8	&40.2&&54.4&	38.5&	11.3&	38.9                        \\ \bottomrule
\end{tabular}

\vspace{4px}
\caption{Robustness testing on Lyft to Ithaca365, localization error. Metrics are reported on nuScenes mAP at 1m matching.}
\vspace{-10px}

\label{tab:robustness-localization-05bev}
\end{table}
\begin{table}[]
\tabcolsep 1pt
\centering
\begin{tabular}{@{}llcccccccc@{}}
\toprule
\# Traversals & \multicolumn{4}{c}{Car}                                                                                     &  & \multicolumn{4}{c}{Pedestrian}                                                                              \\ \cmidrule(lr){2-5} \cmidrule(l){7-10} 
           & \multicolumn{1}{c}{0-30} & \multicolumn{1}{c}{30-50} & \multicolumn{1}{c}{50-80} & \multicolumn{1}{c}{0-80} &  & \multicolumn{1}{c}{0-30} & \multicolumn{1}{c}{30-50} & \multicolumn{1}{c}{50-80} & \multicolumn{1}{c}{0-80} \\ \midrule
$N = 1$      &     56.0&	33.1	&14.0&	36.3&&	52.8&	32.1&	6.3	&33.9                         \\
$N \leq 2$      &   \textbf{59.2}	&39.6&	22.2&	41.9&&	58.4	&43.5&	16.0&	43.8                         \\
$N \leq 3$      &      59.0&	\textbf{39.9}	&24.3&	42.6&&	59.8&	46.0	&19.6	&46.5                         \\
$N \leq 4$      &    59.1	&39.7&	\textbf{25.2}&	\textbf{42.9}&&	\textbf{60.6}&	\textbf{48.4}&	\textbf{20.9}&	\textbf{48.3}                          \\ \bottomrule
\end{tabular}

\vspace{4px}
\caption{Robustness testing on Lyft to Ithaca365, number of traversals. Metrics are reported on nuScenes mAP at 1m matching.}
\label{tab:robustness-traversals-05bev}
\vspace{-10px}

\end{table}

\section{Discussion and Future Works}
\label{sec:discussion}
In this work, we propose our method, \methodshort, for the task of domain adaptation in self-driving object detection. Our work is able to achieve strong performance by training well aligned features from past traversal statistics, and further leverage the statistics to stabilize model outputs in the test domain in an unsupervised manner. Our method is the first to approach domain adaptation for 3D object detection under a feature alignment perspective leveraging past traversal information. Furthermore, by bringing in an architecture specifically designed to leverage such information, we show state-of-the-art performance on two large, real world datasets. Future directions include expanding this framework into other object detectors and exploring other feature alignment methods leveraging past traversals.


{\small
\bibliographystyle{ieee_fullname}
\bibliography{main}
}

\end{document}